# Random Bits Regression: a Strong General Predictor for Big Data

Yi Wang[1,2†], Yi Li[1,2†], Momiao Xiong[1,2,3], Li Jin[1,2*]

[1]State Key Laboratory of Genetic Engineering and Ministry of Education Key Laboratory of Contemporary Anthropology, Collaborative Innovation Center for Genetics and Development and School of Life Sciences, Fudan University, Shanghai, China

[2]Ministry of Education Key Laboratory of Contemporary Anthropology and State Key Laboratory of Genetic Engineering, Collaborative Innovation Center for Genetics and Development, School of Life Sciences, Fudan University, Shanghai, China

[3]Human Genetics Center, School of Public Health, University of Texas Houston Health Sciences Center, Houston, Texas, USA.

**[†]The two authors contributed equally to the work.**

**[*]Corresponding author:**

Dr. Li Jin, School of Life Sciences, Fudan University,

2005 Songhu Road, Shanghai 200433, China

Tel: 8602151630607

Fax: 8602151630607

Email: lijin@fudan.edu.cn

## ABSTRACT

To improve accuracy and speed of regressions and classifications, we present a data-based prediction method, Random Bits Regression (RBR). This method first generates a large number of random binary intermediate/derived features based on the original input matrix, and then performs regularized linear/logistic regression on those intermediate/derived features to predict the outcome. Benchmark analyses on a simulated dataset, UCI machine learning repository datasets and a GWAS dataset showed that RBR outperforms other popular methods in accuracy and robustness. RBR (available on https://sourceforge.net/projects/rbr/) is very fast and requires reasonable memories, therefore, provides a strong, robust and fast predictor in the big data era.

## 1. INTRODUCTION

Data-based modeling is becoming practical in predicting outcomes. We are interested in a general data-based prediction task: given a training data matrix (TrX), a training outcome vector (TrY) and a test data matrix (TeX), predict test outcome vector ($\hat{Y}$). In the era of big data, two practically conflicting challenges are eminent: (1) the prior knowledge on the subject (also known as domain specific knowledge) is largely insufficient; (2) computation and storage cost of big data is unaffordable.

To meet these aforementioned challenges, this paper is devoted to modeling large number of observations without domain specific knowledge, using regression and classification. The methods widely used for regression and classification can be classified as: linear regression, k nearest neighbor(KNN)[1], support vector machine (SVM)[2], neural network (NN)[3, 4], extreme learning machine (ELM)[5], deep learning (DL)[6], random forest (RF)[7] and boosting (GBM)[8] among others. Each method performs well on some types of datasets but has its own limitations on others[9-12]. A method with reasonable performance on boarder, if not universe, datasets is highly desired.

Some prediction approaches (SVM, NN, ELM and DL) share a common characteristics: employing intermediate features. SVM employs fixed kernels as intermediate features centered at each sample. NN and DL learn and tune sigmoid intermediate features. ELM uses a small number (<500) of randomly generated features. Despite their successes, each has its own drawbacks: SVM kernel and its parameters need to be tuned by the user, and the requirement for memory is large: O(sample$^2$). NN and DL's features are learnt and tuned iteratively which is computationally expensive. The number of ELM's features is usually too small for complex tasks. These drawbacks limit their applicabilities on complex tasks, especially when the data is big.

In this report, we propose a novel strategy to take advantage of large number of intermediate features following Cover's theorem[13], which is named Random Bits Regression (RBR). We first generate a huge number of ($10^4$~$10^6$) random intermediate features given TrX, and then utilize TrY to select predictive intermediate features by regularized linear/logistic regression. In order to keep the memory footprint small and compute quickly when employing such huge number of intermediate features, we restrict these features to be binary.

## 2. METHODS

### *2.1 Data Pre-processing*

Suppose that there are $m$ variables $x_1,..., x_m$ as predictors. The data are divided into two parts: training dataset and test dataset. The algorithm takes three input files: TrX, TeX and TrY. TrX and TeX are predictor matrices for the training and test datasets, respectively. Each row represents a sample and each column represents a variable. TrY is a target vector or a response vector, which can have a real valued or binary. We standardize (subtract the mean and divide by the standard deviation) TrX and TeX to ease subsequent processing.

### *2.2 Intermediate Feature Generation*

We generate $10^4$~$10^6$ random binary intermediate features for each sample. Let $K$ be the number of features to be generated and $F = \begin{bmatrix} f_{11} & \cdots & f_{1K} \\ \vdots & \vdots & \vdots \\ f_{n1} & \cdots & f_{nK} \end{bmatrix}$ be the feature matrix where $f_{ij}$ is the *j*th intermediate feature of the *i*th sample. The *k*th intermediate feature vector $f_k = [f_{1k},..., f_{nk}]^T$ is generated as follows:

(1) Randomly select a small subset of variables, e.g. $x_1$, $x_3$, $x_6$.

(2) Randomly assign weights to each selected variables. The weights are sampled from standard normal distribution, for example, $w_1$, $w_3$, $w_6$~N(0,1)

(3) Obtain the weighted sum for each sample, for example $z_i = w_1x_{1i} + w_3x_{3i} + w_6x_{6i}$ for the *i*th sample.

(4) Randomly pick one $z_i$ from the $n$ generated $z_i, i = 1,..., n$ as the threshold T.

(5) Assign bits values to $f_k$ according to the threshold T, $f_{ik} = \begin{cases} 1, z_i \geq T \\ 0, z_i < T \end{cases}, i = 1, \ldots, n$.

The process is repeated K times. The first feature is fixed to 1 to act as the interceptor. The bits are stored in a compact way that is memory efficient (32 times smaller than the real valued counterpart). Once the binary intermediate features matrix F is generated, it is used as the only predictors.

### *2.3 L2 Regularized Linear Regression/Logistic Regression*

For real valued TrY, we apply $L_2$ regularized regression (ridge regression) on F and TrY. We model $\hat{Y}_i = \sum_j \beta_j F_{ij}$, where $\beta$ is the regression coefficient. The loss function to be minimized is $Loss = \sum_i (TrY_i - \hat{Y}_i)^2 + \frac{\lambda}{2} \sum_{j \neq 1} \beta_j^2$, where $\lambda$ is a regularization parameter which can be selected by cross validation or provided by the user. The $\beta$ is estimated by $\hat{\beta} = \arg\min_\beta Loss$.

For binary valued TrY, we apply $L_2$ regularized logistic regression on F and TrY. We model $\hat{Y}_i = \frac{1}{1 + \exp(-\sum_j \beta_j F_{ij})}$, where $\beta$ is the regression coefficient. The loss function to be minimized is $Loss = \sum_i -TrY \ln \hat{Y} - (1 - TrY) \ln(1 - \hat{Y}) + \frac{\lambda}{2} \sum_{j \neq 1} \beta_j^2$, where $\lambda$ is a regularization parameter. The $\beta$ is estimated by $\hat{\beta} = \arg\min_\beta Loss$.

These models are standard statistical models[14]. The L-BFGS (Limited-memory Broyden–Fletcher–Goldfarb–Shanno (BFGS) algorithm) library was employed to perform the parameter estimation. The L-BFGS method only requires the gradient of the loss function and approximates the Hessian matrix with limited memory cost.

Prediction is performed once the model parameters are estimated. Specifically, the same weights that generated the intermediate features in the training dataset were used to generate the intermediate features in the test dataset and use the estimated $\hat{\beta}$ in the training dataset to predict the phenotype Y in the test dataset.

Some optimization techniques are used to speed up the estimation: (1) using a relatively large memory (~1GB) to further speed up the convergence of L-BFGS by a factor of 5, (2) using SSE (Streaming SIMD Extensions) hardware instructions to perform bit-float calculations which speeds up the naive algorithm by a factor of 5, and (3) using multi-core parallelism with OpenMP (Open Multi-Processing) to speed up the algorithm.

### *2.4 Benchmarking*

We benchmarked nine methods including linear regression (Linear), logistic regression (LR), k-nearest neighbor (KNN), neural network (NN), support vector machine (SVM), extreme learning machine (ELM), random forest (RF), generalized boosted regression models (GBM) and random bit regression (RBR). Our RBR method and usage are available on the website (https://sourceforge.net/projects/rbr/). The KNN method was implemented by our own C++ code. The other seven methods were implemented by R (version: 3.0.2) package: stats, nnet (version: 7.3-8), kernlab (version: 0.9-19), randomForest (version: 4.6-10), elmNN (version: 1.0), gbm (version: 2.1) accordingly. Ten-fold cross validation was used to evaluate their performance. For methods that are sensitive to parameters, the parameters were manually tuned to obtain the best performances. The benchmarking was performed on a desktop PC, equipped with an AMD FX-8320 CPU and 32GB memory. The SVM on some large sample datasets failed to finish the benchmarking within a reasonable time (2 week). Those results are left as blank.

We first benchmarked all methods on a simulated dataset. The dataset contains 1,000 training samples and 1,000 testing samples. It contains two variables (X, Y) and is created with the simple formula: $Y = Sin(X) + N(0,0.1), X \in (-10\pi, 10\pi)$.

We then benchmarked all datasets from the UCI machine learning repository[15] with the following inclusion criterion: (1) the dataset contains no missing values; (2) the dataset is in dense matrix form; (3) for classification, only binary classification datasets are included; and (4) the included dataset should have a clear instruction and the target variable should be specified.

Overall, we tested 14 regression datasets. They are: 1) 3D Road Network[16], 2)Bike sharing[17], 3) buzz in social media tomhardware, 4) buzz in social media twitter, 5) computer hardware[18], 6) concrete compressive strength[19], 7) forest fire[20], 8) Housing[21], 9) istanbul stock exchange[22], 10) parkinsons telemonitoring[23], 11) Physicochemical properties of protein tertiary structure, 12) wine quality[24], 13) yacht hydrodynamics[25], and 14) year prediction MSD[26]. In addition, we tested 15 classification datasets: 1) banknote authentication, 2) blood transfusion service center[27], 3) breast cancer wisconsin diagnostic[28], 4) climate model simulation crashes[29], 5) connectionist bench[30], 6) EEG eye state, 7) fertility[31], 8) habermans survival[32], 9) hill valley with noise[33], 10) hill valley without noise[33], 11)Indian liver patient[34], 12) ionosphere[35], 13) MAGIC gamma telescope[36], 14) QSAR biodegradation[37], and 15) skin segmentation[38].

All methods were also applied on one psoriasis[39] GWAS genetic dataset to predict disease outcomes. We used a SNP ranking method for feature selection which was based on allelic association p-values in the training datasets, and selected top associated SNPs as input variables. To ensure the SNP genotyping quality, we removed SNPs that were not in HWE (Hardy-Weinberg Equilibrium) (p-value < 0.01) in the control population.

## 3. RESULTS

We first examined the nonlinear approximation accuracy of the 8 methods. Figure1 shows the curve fitting for the sine function with several learning algorithms. We observed that linear regression, ELM and GBM failed on this dataset and the SVM's fitting was also not satisfactory. On the contrary, KNN, NN, RF and RBR produced good results.

Next we evaluated the performance of the eight methods for regression analysis. Table 1 showed the average regression RMSE (root-mean-square error) of the eight methods on 14 datasets (see detailed description of databases). We observed several remarkable features from Table 1. First, the RBR took 10 first places, 3 second places and 1 third places among the 14 datasets. In the cases that RBR was not in first place, the difference between the RBR and the best prediction was within 2%. RBR did not experience any breakdown for all 14 datasets. The random forest was the second best method, however, it suffered from failure on the ***yacht hydrodynamics*** dataset.

Finally, we investigated the performance of the RBR for classification. Table 2 showed the classification error percentages of different methods on 16 datasets. RBR took 12 first places, and 4 second places. In the cases when the RBR was not the first place, the difference between the RBR method and the best classification was small and no failure was observed. Despite its simplicity, KNN was the second best method and took 3 first places. However, it suffered from failure/breakdown on the ***Climate Model Simulation Crashes***, ***EEG Eye State***, ***Hill Valley with noise***, ***Hill Valley without noise***, and the ***Ionosphere*** dataset.

The RBR is also reasonably fast on big datasets. For example, it took two hours to process the largest dataset ***year prediction MSD*** (515,345 samples, 90 features, and $10^5$ intermediate features).

## 4. DISCUSSIONS

Big data analysis consists of three scenarios: (1) a large number of observations with limited number of features, (2) a large number of features with limited number of observations and (3) both numbers of observations and features are large. This paper focuses on the large number of observations with limited number of features. We have addressed three key issues for big observation data analysis.

The first issue is how to split the sample space into sub-sample space. The RBR has geometric interpretations: each intermediate feature (bit) split the sample space into two parts and serves a basis function for regression. In one dimensional cases, it approximates functions by a set of weighted step functions. In two dimensional cases, the large number of bits split the plane into mosaic-like regions. By assigning corresponding weight to each bit, these regions can approximate 2-D functions. For high dimensional spaces, the interpretation is similar to 2-D cases. Therefore, the RBR method with a large number of intermediate features split the whole sample space into many relatively homogeneous sub-sample spaces. The RBR is similar to ELM, especially the one proposed by Huang et al.[40]. The differences between them are (1) the amount of intermediate features used, (2) the random feature generation and (3) the optimization. The RBR utilizes a huge number of features ($10^4$~$10^6$) and the ELM uses a much small number (<500). The ELM is small due to two reasons: (1) computational cost: O(intermediate feature$^3$). (2) accuracy problem. In the ELM larger number of features does not always lead to better prediction, usually ~100 features is the best choice. On the contrary, the RBR's computational cost is O(intermediate feature) and a larger number of features usually leads to better precision due to regularization. RBR's random feature generation differs from that of the ELM. The choice of sample based threshold ensures that the random bit divides the sample space uniformly; on the contrary the ELM's random feature does not guarantee uniform partition of the samples. It tends to focus the hidden units on the center of the dataset thus badly fitting the remainder of the sample space (Figure 1). The L-BFGS and SSE optimization and multi-core parallelism make RBR 100 times faster than the ELM when the same number of feature is employed. Huang et al. provide some theoretical results for both the RBR and ELM.

The second issue is how the results from each of the subsets are then combined to obtain an overall result. The RBR is closely related to boosting. Each RBR random bit can be viewed as a weak classifier. Logistic regression is the same as one kind of boosting algorithm named logit-boost. The RBR method boosts those weak bits to form a strong classifier. The RBR is closely related to neural networks. The RBR is equivalent to a single hidden layer neural network and the bits are the hidden units. Large number of bits is a conjugate fashion (we call it **wide learning**) to deep learning. As no back-propagation is required, the learning rule is quite simple, thus is biologically feasible. Biologically, the brain has the capacity to form a huge feature layer (maybe $10^8$~$10^{10}$) to approximates functions well.

The third issue is computational cost. The RBR scales well in memory and computation time compared to the SVM due to a fixed number of binary features. The RBR is faster than the random forest or boosting trees due to the light weight nature of the bits.

## 5. CONCLUSION

In conclusion, we can confidently conclude that the RBR is a strong, robust and fast off-the-shelf predictor especially in the big data era.

## CONFLICT OF INTEREST

There are no conflicts of interest.

## ACKNOWLEDGMENTS

There are no funding support.

**Figure 1. Fitting a sine curve.**

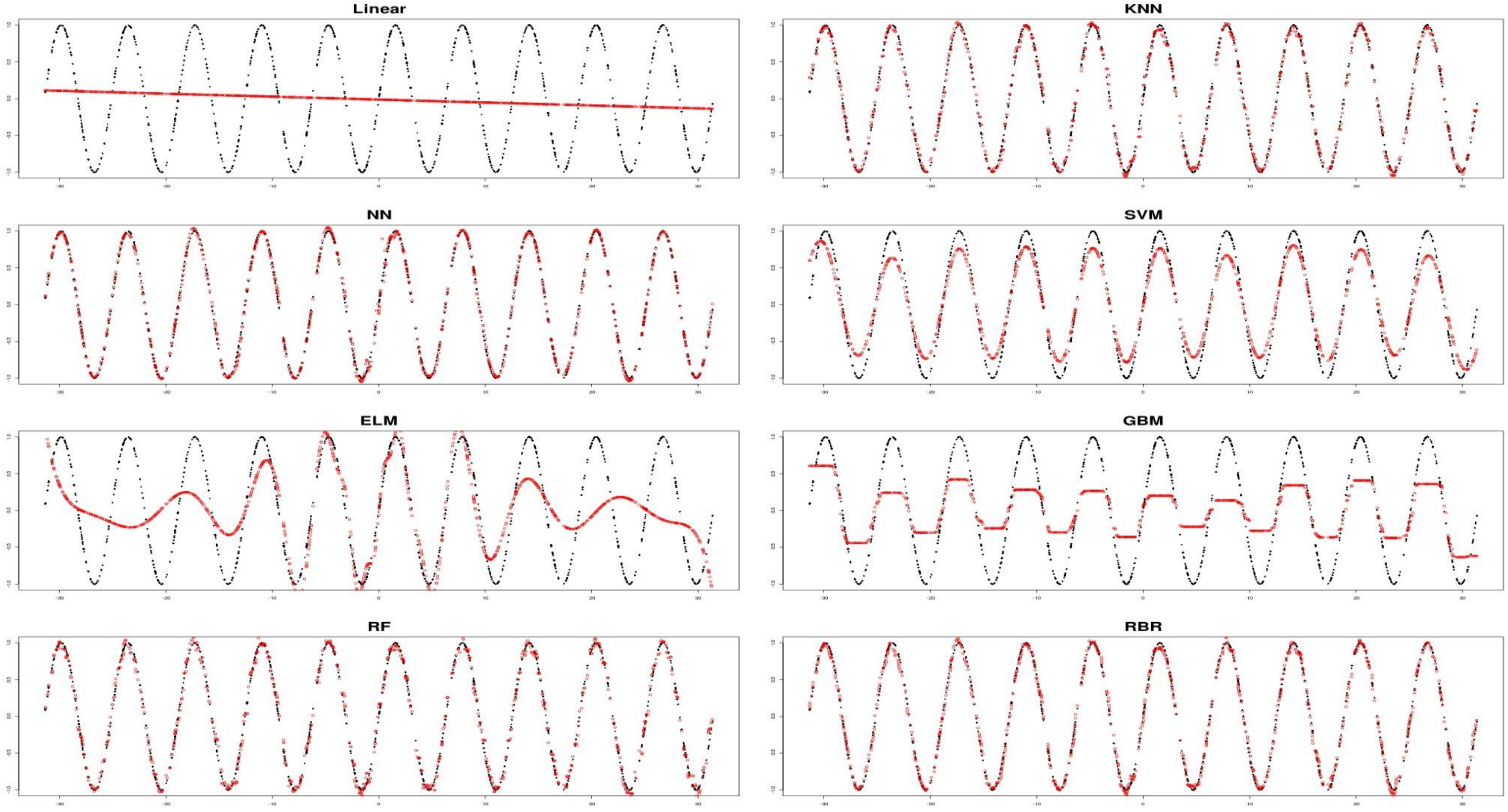

Black dots are the theoretical values while red dots are fitted values.

**Table 1.** Regression RMSE of different methods

| RMSE | Sample | Feature | Linear | KNN | NN | SVM | ELM | GBM | RF | RBR |
|---|---|---|---|---|---|---|---|---|---|---|
| **3D Road Network** | 434874 | 2 | 18.370 | 6.441 | 15.548 | 12.530 | 16.953 | 14.819 | 3.855 | **2.061** |
| **Bike_sharing** | 17389 | 16 | 141.865 | 104.576 | 65.994 | 114.155 | 94.564 | 96.765 | 49.366 | **40.54** |
| **buzz_in_social_media_tomhardware*** | 28179 | 97 | 1.446 | 0.758 | 0.373 | 1.489 | 1.581 | 0.311 | **0.310** | 0.313 |
| **buzz_in_social_media_twitter*** | 583250 | 78 | 1.333 | 0.516 | 0.505 | - | 1.034 | 0.484 | **0.471** | 0.472 |
| **computer_hardware** | 209 | 7 | 69.622 | 63.125 | 134.912 | 119.394 | 159.233 | 93.214 | 61.212 | **50.001** |
| **concrete_compressive_strength** | 1030 | 9 | 10.530 | 8.280 | 6.355 | 6.519 | 13.176 | 5.823 | 5.096 | **3.650** |
| **forest_fire*** | 517 | 13 | 1.503 | 1.399 | 2.095 | 1.499 | 1.401 | 1.399 | 1.454 | **1.390** |
| **Housing** | 506 | 12 | 4.884 | 4.099 | 4.943 | 3.752 | 7.922 | 3.749 | 3.097 | **2.770** |
| **istanbul_stock_exchange** | 536 | 8 | **0.012** | 0.013 | 0.039 | 0.013 | 0.016 | **0.012** | 0.013 | **0.012** |
| **parkinsons_telemonitoring** | 5875 | 26 | 9.741 | 6.097 | 6.690 | 7.160 | 10.354 | 6.889 | **3.909** | 3.954 |
| **Physicochemical_properties_of _protein_tertiary_structure** | 45730 | 9 | 5.185 | 3.790 | 6.118 | 6.254 | 6.118 | 5.047 | 3.454 | **3.407** |
| **wine_quality** | 6497 | 11 | 0.736 | 0.696 | 0.730 | 0.676 | 0.921 | 0.701 | **0.585** | 0.592 |
| **yacht_hydrodynamics** | 308 | 6 | 9.134 | 6.430 | 1.178 | 6.542 | 1.964 | 1.160 | 3.833 | **0.782** |
| **year_prediction_MSD** | 515345 | 90 | 9.550 | 9.216 | 10.931 | - | 11.468 | 9.626 | 9.242 | **9.144** |

The * means the dependent variable of the corresponding data was transformed by log function to be more asymptotically normal.

The bold means the first place result of all methods compared.

**Table 2.** Classification error rates of difference methods

| error% | Sample | Feature | LR | KNN | NN | SVM | ELM | GBM | RF | RBR |
|---|---|---|---|---|---|---|---|---|---|---|
| **banknote_authentication** | 1372 | 4 | 1.018 | 0.146 | **0.000** | **0.000** | **0.000** | 0.801 | 0.656 | **0.000** |
| **Blood_Transfusion_Service_Center** | 748 | 4 | 22.863 | 19.649 | 24.458 | 20.186 | 23.802 | 23.667 | 24.596 | **19.521** |
| **Breast_Cancer_Wisconsin_Diagnostic** | 569 | 30 | 5.091 | 2.810 | 8.446 | 2.456 | 8.800 | 3.863 | 4.211 | **2.281** |
| **Climate_Model_Simulation_Crashes** | 540 | 18 | 4.259 | 7.037 | 5.556 | 7.778 | 5.926 | 6.296 | 7.593 | **3.888** |
| **Connectionist_Bench** | 208 | 60 | 26.000 | 13.023 | 21.667 | 13.476 | 14.429 | 16.833 | 13.452 | **11.571** |
| **EEG_Eye_State** | 14980 | 14 | 35.748 | 15.374 | 31.569 | 19.519 | 42.336 | 24.172 | **6.001** | 6.612 |
| **Fertility** | 100 | 9 | 15.000 | **12.000** | 15.000 | **12.000** | 24.000 | **12.000** | 14.000 | **12.000** |
| **habermans_survival** | 306 | 3 | 25.849 | 25.160 | 30.710 | 26.742 | 27.400 | 27.774 | 27.409 | **25.118** |
| **Hill_Valley_with_noise** | 1212 | 100 | 42.001 | 45.707 | 5.280 | 46.283 | 23.422 | 50.906 | 43.065 | **4.041** |
| **Hill_Valley_without_noise** | 1212 | 100 | 41.340 | 41.668 | **0.000** | 46.618 | 15.596 | 51.734 | 39.602 | 0.744 |
| **Indian_Liver_Patient** | 579 | 10 | 27.828 | 27.822 | 30.206 | 28.684 | 28.336 | 28.336 | 29.189 | **27.644** |
| **Ionosphere** | 351 | 34 | 10.262 | 10.246 | 11.984 | **5.405** | 10.278 | 6.825 | 7.405 | 5.413 |
| **MAGIC_Gamma_Telescope** | 19020 | 10 | 20.878 | 15.857 | 13.170 | 12.976 | 22.639 | 13.991 | 11.725 | **11.435** |
| **QSAR_biodegradation** | 1055 | 41 | 13.366 | 13.754 | 14.978 | 12.144 | 22.381 | 14.884 | 13.180 | **12.043** |
| **Skin_Segmentation** | 245057 | 3 | 8.121 | 0.040 | 0.056 | 0.081 | 0.263 | 1.550 | 0.043 | **0.039** |
| **Psoriasis** | 1590 | 68-88 | 40.566 | **37.044** | 42.327 | 38.176 | 38.616 | 40.818 | 40.440 | 37.170 |

The bold means the first place result of all methods compared.